\documentclass[runningheads]{llncs}
\usepackage[T1]{fontenc}
\usepackage{graphicx,verbatim}
\usepackage{booktabs}
\usepackage{multirow}
\usepackage{makecell}
\usepackage{amsmath}
\usepackage{amssymb}
\usepackage{xcolor}
\usepackage[table]{xcolor}  
\usepackage{colortbl}
\usepackage{booktabs}

\newcommand{\yichen}[1]{{\color{black}#1}}

\begin{document}
%
\title{MAPLE: Elevating Medical Reasoning from Statistical Consensus to Process-Led Alignment}
\titlerunning{MAPLE}
%

\author{Kailong Fan\inst{1,3}\thanks{Equal contribution.} \and
Anqi Pu\inst{2} \protect\footnotemark[1]\and
Yichen Wu\inst{3}$^{\dagger}$ \and
Wanhua Li\inst{2} \and
Yicong Li\inst{2} \and
Hanspeter Pfister\inst{2} \and
Huafeng Liu\inst{1} \and
Xiang Li\inst{3} \and
Quanzheng Li\inst{3} \and
Ning Guo\inst{3}}  
\authorrunning{Fan, Pu et al.}
\institute{$^{1}$\,Zhejiang University, 
$^{2}$\,Harvard University, 
$^{3}$\,Harvard Medical School \\
$^{\dagger}$Corresponding author: yiwu6@mgh.harvard.edu}
  
\maketitle              
\begin{abstract}
\yichen{Recent advances in medical large language models have explored Test-Time Reinforcement Learning (TTRL) to enhance reasoning. However, standard TTRL often relies on majority voting (MV) as a heuristic supervision signal, which can be unreliable in complex medical scenarios where the most frequent reasoning path is not necessarily the clinically correct one. In this work, we propose a novel and unified training paradigm that integrates medical process reward models with TTRL to bridge the gap between test-time scaling (TTS) and parametric model optimization. Specifically, we advance the TTRL framework by replacing the conventional MV with a fine-grained, expert-aligned supervision paradigm using Med-RPM. This integration ensures that reinforcement learning is guided by medical correctness rather than mere consensus, effectively distilling search-based intelligence into the model's parametric memory. Extensive evaluations on four different benchmarks have demonstrated that our developed method consistently and significantly outperforms current TTRL and standalone PRM selection. Our findings establish that transitioning from stochastic heuristics to structured, step-wise rewards is essential for developing reliable and scalable medical AI systems.}



\end{abstract}
\section{Introduction}
Large language models (LLMs) are increasingly \yichen{studied for medical decision support, including radiology interpretation ~\cite{Bhayana2024Chatbots,Nam2025Multimodal,Zhang2025Multimodal}, clinical question answering~\cite{Singhal2025Toward,Wang2024Large}, and multi-step diagnostic reasoning ~\cite{Dinc2025Comparative,Goh2024Large}. In these safety-critical settings, errors are not merely utility-degrading but can translate into clinically inappropriate decisions~\cite{Kim2025Limitations,Hager2024Evaluation}, so improving reasoning reliability is a primary requirement rather than an optional enhancement. Medical reasoning further differs from many general-domain tasks because correct answers often depend on a sequence of clinically grounded intermediate judgements, where early mistakes can cascade and become difficult to recover from.}


\yichen{A common approach to improve reliability is TTS~\cite{Zhang2025A,Ji2025A}, where the model samples multiple reasoning trajectories and aggregates them using self-consistency, typically implemented as MV. This strategy can reduce variance when errors are mostly independent, but it has a structural limitation: frequency is not a proxy for clinical correctness when the sampled trajectories share correlated misconceptions or systematically omit key evidence. In medical problems, such a correlation is plausible because many trajectories are produced by the same model, share the same bind spots, and can converge to the same incorrect but internally coherent explanation ~\cite{Gu2025MedAgentAudit:}. As a result, the most frequent reasoning path can still be wrong even when it appears consistent. Recent works therefore move from outcome-only aggregation to process-level verification ~\cite{Lightman2023Let's,Wang2024Multi-step}, using external or auxiliary verifiers to judge intermediate steps. Medical process reward models, such as Med-PRM~\cite{Yun2025Med-PRM}, provide a clinically aligned signal by evaluating the correctness of intermediate reasoning and using these scores to rerank candidate trajectories.}

\yichen{However, verification-based methods are largely selection-only ~\cite{Zhang2024Generative,Wang2022Self-Consistency}. They improve final answers by choosing the best candidate from a sampled pool, while leaving the underlying generator unchanged. This creates two practical constraints. First, performance gains must be repeatedly paid for with sampling and reranking at inference time, which limits scalability under tight latency or cost budgets. Second, systematic errors can persist because the model's proposal distribution is not corrected; the verifier can filter mistakes but cannot prevent the generator from repeatedly producing them. In parallel, TTRL has been proposed as a mechanism to improve the generator directly on unlabeled test inputs ~\cite{Zuo2025TTRL,Liu2025ETTRL:,Zhang2025AQA-TTRL:}. The key idea is to define a proxy supervision signal from the model's own samples, then perform small policy updates so that future generations are more likely to match that signal.}


\yichen{This paper argues that the missing piece is a medically grounded supervision signal that can replace voting inside TTRL. If process verifiers can reliably distinguish clinically correct reasoning from plausible but incorrect reasoning at test time, then their preferences should be used not only to select outputs but also to guide policy improvement. Motivated by this observation, we propose \textbf{MAPLE} (\textbf{M}edical \textbf{A}lignment via \textbf{P}rocess-\textbf{L}ed \textbf{E}volution), a test-time training paradigm that integrates medical process reward models with TTRL to bridge TTS and parametric optimization. Concretely, MAPLE replaces MV pseudo-supervision with fine-grained, expert-aligned step-wise rewards produced by a medical process reward model (Med-RPM). Instead of optimizing toward what the model most frequently says, MAPLE optimizes toward what the medical verifier judges as correct along the reasoning process, thereby aligning test-time learning with clinical validity rather than consensus. In summary, our main contributions are: }
\vspace{-1.8mm}
\begin{itemize}
    \item We introduce a unified paradigm that bridges TTS and TTRL, enabling generate-and-improve on unlabeled medical queries.
    \item We propose MAPLE, which replaces vote-based supervision in TTRL with step-wise rewards to guide medically grounded test-time updates. 
    \item We conduct extensive experiments on four medical reasoning benchmarks, demonstrating consistent improvements over vote-based TTRL and PRM-only test-time selection, with additional studies confirming the benefit of process-led rewards for stable and effective test-time adaptation.
\end{itemize}

\section{Background}
\noindent \textbf{TTS and Self-Evolution.} \yichen{Increasing computation at inference time has become a practical way to elicit stronger reasoning from LLMs, often complementing or even rivaling gains from scaling model size~\cite{Snell2024Scaling,Huang2025m1}. A dominant line of work samples multiple solutions and aggregates them via consensus (e.g., Self-Consistency) or selection (e.g., Best-of-M with a reward model)~\cite{Wang2022Self-Consistency,Snell2024Scaling}. Beyond selection-only schemes, TTRL enables parameter adaptation on unlabeled inputs by deriving proxy rewards from repeated sampling and MV, yielding \textit{self-evolution} at test time~\cite{Zuo2025TTRL}. While recent frontier reasoning models leverage outcome-verifiable rewards or expensive human annotations for RL~\cite{Guo2025DeepSeek-R1,El-Kishky2024OpenAI}, unlabeled test-time adaptation is particularly appealing in medical settings where real-time gold standards are rarely available~\cite{Hager2024Evaluation}. }

\vspace{0.6em}
\noindent \textbf{Inference-Time Verification: From Outcome to Process Rewards.} \yichen{Test-Time Scaling improves output quality without updating model parameters, typically by exploring multiple reasoning trajectories and selecting the most reliable one. Outcome-based reward models score whole solutions but can misjudge trajectories due to spurious reasoning that happens to reach the correct answer or correct early steps that later derail~\cite{Yuan2025Curing,Lee2025Rethinking}. Process reward models address this by providing stepwise supervision to evaluate intermediate reasoning validity~\cite{Lightman2023Let's,Zheng2025A}. In medicine, stepwise verification is challenging due to the lack of symbolic verifiers; Med-PRM tackles this by grounding step evaluations in retrieved clinical guidelines and medical literature (RAG-as-a-judge), enabling stronger verifier-based selection in medical reasoning~\cite{Yun2025Med-PRM,Liu2025Improving}.}

\begin{figure}[t]
  \centering
  \includegraphics[width=\linewidth]{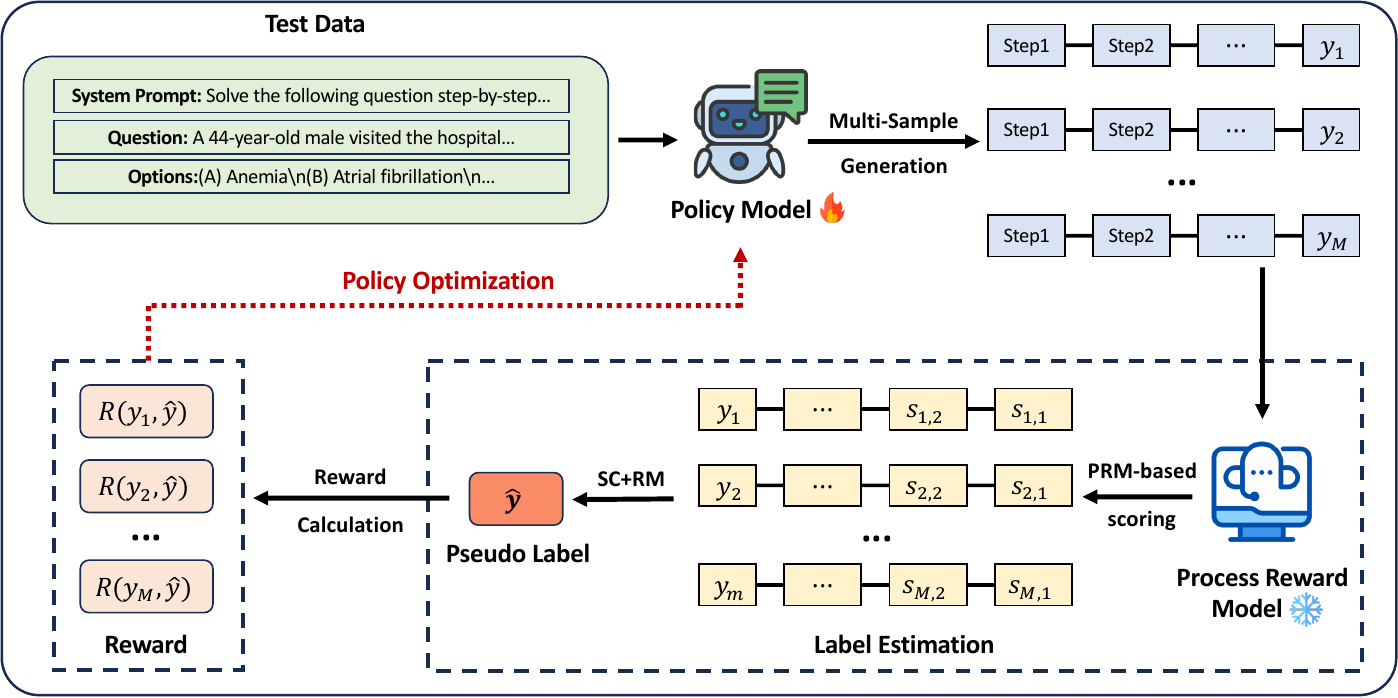}
  \caption{
Overview of the MAPLE framework. Given a test question, the policy 
model generates $M$ candidate reasoning chains via multi-sample generation. Each chain is evaluated by a PRM, which assigns step-level scores $s_{i,j}$ to every intermediate reasoning step. The scored candidates are aggregated through a Self-Consistency with Reward Model reranking (SC+RM) mechanism to produce a pseudo label $\hat{y}$. Per-sampel rewards $R(y_i, \hat{y})$
are then computed by the candidate and pseudo label, and the resulting reward signals are used to update the policy model online via policy optimization.}
  \label{fig:medttrl}
\end{figure}

\section{\yichen{The Proposed MAPLE Method}}
\yichen{
We propose the MAPLE algorithm, a PRM-guided TTRL framework for medical reasoning. Concretely, it combines process-level verification with test-time optimization, so the model can learn from the verifier feedback at inference rather than relying only on post-hoc reranking.

\vspace{-0.6mm}
\subsection{Preliminary}
Given a medical question $x$, a policy model $\pi_{\theta}$ samples $M$ reasoning trajectories
\begin{equation}
\{y_i\}_{i=1}^{M}, \qquad y_i \sim \pi_{\theta}(\cdot \mid x),
\end{equation}
where each trajectory $y_i$ contains a step-by-step rationale followed by a final predicted answer $a_i \in \mathcal{A}$.
Following prior work on medical process verification~\cite{Yun2025Med-PRM}, we constrain the number of reasoning steps to a moderate range ($2 < T_i < 10$) to discourage degenerate or overly shallow rationales.

\vspace{0.6mm}
\noindent \textbf{Process reward model.}
We employ a medical process reward model that evaluates the intermediate steps of each trajectory and outputs step-level scores,
\begin{equation}
\{s_{i,t}\}_{t=1}^{T_i}, \qquad 2 < T_i < 10,
\end{equation}
where larger $s_{i,t}$ indicates higher medical correctness and logical validity at step $t$.
To obtain a conservative trajectory-level confidence, we aggregate step scores via a worst-step rule:
\begin{equation}
S_i = \min_{t} s_{i,t}.
\end{equation}
This design reflects the safety-critical nature of medical reasoning, where a single incorrect intermediate step may invalidate the final conclusion.

\subsection{MAPLE: PRM-Guided Test-Time Reinforcement Learning}
As shown in Fig.~\ref{fig:medttrl}, the proposed MAPLE method is a PRM-guided ttrl framework to improve medical reasoning without additional supervised annotations.
At a high level, MAPLE consists of three stages executed at test time for each input $x$:
(1) sample multiple trajectories from $\pi_{\theta}$ and score them with the PRM;
(2) estimate a pseudo label $\hat{a}$ using PRM-guided self-consistency;
(3) update $\pi_{\theta}$ with a reinforcement learning objective that encourages trajectories consistent with $\hat{a}$.
This procedure distills verifier-guided selection signals into the model parameters, turning repeated sampling into persistent policy improvement.

\noindent \textbf{Label Estimation via PRM.}
Given $\{(y_i, a_i, S_i)\}_{i=1}^{M}$, we first map PRM confidences to soft weights,
\begin{equation}
w_i = \sigma\!\left(\alpha (S_i - \beta)\right)
= \frac{1}{1 + \exp\!\left(-\alpha (S_i - \beta)\right)},
\label{eq:sigmoid_mapping}
\end{equation}
where $\alpha>0$ controls sharpness and $\beta$ sets the midpoint.
This mapping suppresses low-confidence trajectories while retaining graded preferences among high-confidence candidates, leading to more stable pseudo-label estimation.

We then group trajectories by their predicted answers and compute an aggregated confidence for each answer:
\begin{equation}
R(a)=\sum_{i\in\mathcal{G}(a)} w_i, \qquad
\mathcal{G}(a) = \{i \mid a_i = a\}.
\label{eq:sc_rm}
\end{equation}
The pseudo-label is selected as
\begin{equation}
\hat{a} = \arg\max_{a \in \mathcal{A}} R(a).
\label{eq:pseudo_label}
\end{equation}
When all $w_i$ are equal, this reduces to standard MV; otherwise, it strategically prioritizes answers supported by higher-quality reasoning trajectories that exhibit superior clinical logical consistency as judged by the PRM.

\vspace{1mm}
\noindent \textbf{Reward and TTRL Update.}
Given the pseudo label $\hat{a}$, we define a trajectory-level reward based on answer agreement:
\begin{equation}
r_i =
\begin{cases}
1, & \text{if } a_i = \hat{a}, \\
0, & \text{otherwise}.
\end{cases}
\label{eq:reward}
\end{equation}
Intuitively, PRM-guided self-consistency provides a medically grounded pseudo-supervision target $\hat{a}$, and reinforcement learning internalizes this target by shifting probability mass toward trajectories that yield $\hat{a}$.

We optimize the expected reward under $\pi_{\theta}$:
\begin{equation}
\max_{\theta} \ \mathbb{E}_{y \sim \pi_{\theta}(\cdot \mid x)} \left[ r(y; \hat{a}) \right],
\end{equation}
and update $\theta$ using GRPO~\cite{shao2024deepseekmath} with a small learning rate to ensure conservative test-time adaptation:
\begin{equation}
\theta \leftarrow \theta + \eta \,\nabla_{\theta}\,
\mathbb{E}_{y \sim \pi_{\theta}(\cdot \mid x)} \left[ r(y; \hat{a}) \right].
\end{equation}
}

\section{Experimental Results}
\subsection{Experiment Settings}
\textbf{Baselines.} \yichen{We compare against three complementary baseline categories: (i) strong LLMs
including Llama-3.1-8B-Instruct as well as larger reasoning-oriented models (DeepSeek-R1, QwQ-32B) and a medical-specialized model (HuatuoGPT-o1); (ii) test-time inference scaling baselines represented by Med-PRM~\cite{Yun2025Med-PRM}, which improves performance via PRM-guided reranking/aggregation without updating model parameters; and (iii) test-time reinforcement learning baselines represented by TTRL~\cite{Zuo2025TTRL}, which performs unlabeled test-time policy updates using proxy rewards derived from repeated sampling.}

\vspace{0.6mm}
\noindent \textbf{Benchmark.} \yichen{We evaluate medical reasoning and diagnostic capability on four representative benchmarks: 
\textbf{1) MedQA.} Professional-level multiple-choice questions sourced from the USMLE, requiring multi-step clinical reasoning. We report results on the English 4-option and 5-option subsets.
\textbf{2) MedMCQA.} Large-scale multi-subject medical QA from Indian entrance exams, covering 21 subjects and thousands of topics to assess both factual knowledge and clinical reasoning.
\textbf{3) DDXPlus.} Differential diagnosis benchmark constructed from patient antecedents and structured symptoms (categorical/multi-choice with hierarchical organization), requiring identification of the correct pathology from candidate diagnoses.
\textbf{4) MMLU-Med (Medical Subset).} Medical categories from MMLU
in a standardized multiple-choice format.}

\vspace{0.6mm}
\noindent \textbf{Evaluation Protocol.}
We adopt multiple decoding and selection strategies for evaluation. 
Pass@1 measures the accuracy of a single sampled response under non-zero temperature sampling. 
To evaluate test-time scaling behavior, we consider three inference-time strategies: MV selects the most frequent answer among $M$ sampled responses; Best-of-M (BoM) chooses the candidate with the highest PRM score; and SC+RM groups responses by final answer and selects the group with the highest summed PRM score. 

\subsection{ Experimental Analysis}
\begin{table*}[t]
\centering
\caption{Performance comparison on medical QA benchmarks. Results are reported on MedQA (USMLE-style), MedMCQA, DDXPlus, and MMLU-Med.}
\label{tab:main}
\setlength{\tabcolsep}{6pt}
\begin{tabular}{@{} l c c c c c @{}}
\toprule
Model & Size & MedQA & MedMCQA & DDXPlus & MMLU-Med \\
\midrule
Llama3.1 (pass@1)          & 8B  & 62.79 & 57.44 & 66.41 & 78.33 \\
Llama3.1 (MV)              & 8B  & 68.25 & 60.00 & 74.00 & 83.33 \\
Gemma 2 (pass@1)           & 9B  & 52.90 & 50.97 & 63.80 & 79.63 \\
Gemma 2  (MV)              & 9B  & 54.76 & 52.00 & 69.00 & 79.63 \\
R1-Distill-Llama(pass@1)   & 8B  & 47.10 & 42.63 & 49.22 & 70.25 \\
R1-Distill-Llama (MV)      & 8B  & 61.90 & 44.00 & 80.00 & 78.70 \\
R1-Distill-Qwen (pass@1)   & 7B  & 18.20 & 29.81 & 28.90 & 38.95 \\
R1-Distill-Qwen (MV)       & 7B  & 34.13 & 38.00 & 62.00 & 59.26 \\
HuatuoGPT-o1 (pass@1)      & 8B  & 65.89 & 54.93 & 58.33 & 72.87 \\
HuatuoGPT-o1 (MV)          & 8B  & 65.11 & 52.83 & 58.33 & 74.00 \\
UltraMedical (pass@1)       & 8B  & 57.53 & 50.61 & 63.59 & 66.72 \\
UltraMedical (MV)           & 8B  & 68.25 & 64.00 & 72.00 & 81.48 \\
Med-PRM (BoM)               & 8B  & 71.43 & 63.00 & 79.00 & 78.70 \\
Med-PRM (SC+RM)             & 8B  & 69.84 & 62.00 & 78.00 & 83.33 \\
\rowcolor{gray!15}
MAPLE (BoM)                 & 8B  & \textbf{73.02} & \textbf{66.00} & \textbf{83.00} & \textbf{85.19} \\
\bottomrule
\end{tabular}
\end{table*}

\begin{figure}
    \centering
    \includegraphics[width=1.0\linewidth]{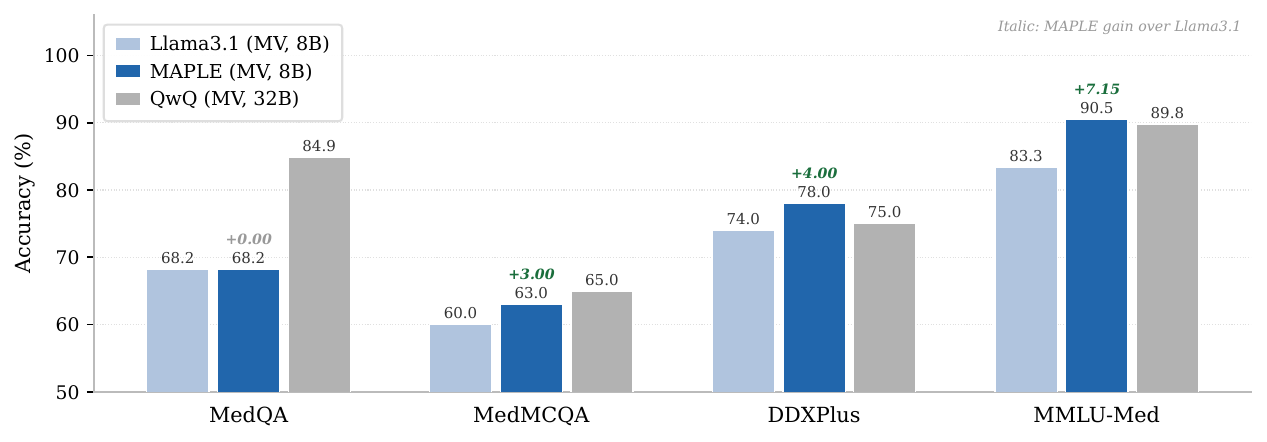}
    \caption{
Performance comparison across four medical QA benchmarks. Built on Llama3.1 (8B) as backbone, MAPLE consistently outperforms its base model and surpasses QwQ (32B) on DDXPlus and MMLU-Med despite being 4× smaller in model size. Italic green values indicate MAPLE's absolute accuracy gain over the Llama3.1 (MV) backbone.
    }
    \label{fig:bar}
\end{figure}

\subsubsection{Main Results.} Table \ref{tab:main} and Fig.~\ref{fig:bar} present a comprehensive comparison of MAPLE against representative baselines across four medical QA benchmarks. By leveraging BOM,
MAPLE achieves SOTA performance among all 8B models, obtaining 73.02\% on MedQA, 66.00\% on MedMCQA, and 83.00\% on DDXPlus.

\vspace{0.6mm}
\noindent MAPLE demonstrates consistent advantages across four comparison axes: \textbf{i) Backbone baselines.} it exceeds its backbone Llama3.1 (MV) by 4.77\%, 6.00\%, 9.00\%, and 1.86\% on the four benchmarks, showing that Med-PRM-guided test-time policy optimization delivers meaningful gains without additional training data, with Fig.~\ref{fig:bar} further indicating larger improvements on MedMCQA, DDXPlus, and MMLU-Med under MV where structured step-by-step reasoning is rewarded; \textbf{ii) Reasoning-distilled models.} at the same 8B scale, MAPLE substantially outperforms R1-Distill-Llama (MV) and surpasses HuatuoGPT-o1 (MV) on all benchmarks, while the sharp drop of domain-agnostic distillation (e.g., R1-Distill-Qwen variants reaching only 18.20\% on MedQA) underscores a risk that MAPLE avoids via test-time adaptation; \textbf{iii) PRM-based methods.} compared to the most direct baseline Med-PRM (BoM), MAPLE achieves further gains of 1.59\%, 3.00\%, 4.00\%, and 6.49\%, suggesting its benefit comes not only from improved candidate selection but from online policy updates that enhance generated reasoning chains beyond what static reranking can exploit; and \textbf{iv) Larger models.} despite being 4$\times$ smaller, MAPLE (8B) surpasses QwQ (32B) on DDXPlus (78.0\% vs.\ 75.0\%) and MMLU-Med (90.5\% vs.\ 89.8\%), highlighting the parameter efficiency of scaling test-time compute.

\subsection{Ablation Study}

\begin{table*}[t]
\centering
\caption{Ablation study on model components across four benchmarks. TTRL: base policy optimized without any PRM guidance; MAPLE-Med$S^3$: TTRL augmented with Med$S^3$ as the process reward model.}
\label{tab:ablation}
\begin{tabular*}{\columnwidth}{@{\extracolsep{\fill}} l c c c c c}
\toprule
Model & Size & 
MedQA & MedMCQA & DDXPlus 
& MMLU-Med \\

\midrule
TTRL                      & 8B & 69.84 & 62.00 & 73.00 & 85.71 \\
TTRL-Med$S^3$ (MV)        & 8B & 69.05 & 59.87 & 74.99 & 80.95\\
TTRL-Med$S^3$ (BoM) & 8B & 69.05 & 61.00 & 78.00 & 82.41 \\
TTRL-Med$S^3$ (SC+RM)     & 8B & 69.05 & 60.00 & 75.00 & 82.41 \\
MAPLE (MV)         & 8B & 68.25 & 63.00 & 78.00 & \textbf{90.48} \\
MAPLE (BoM)  & 8B & \textbf{73.02} & \textbf{66.00} & \textbf{83.00} & 85.19\\
MAPLE (SC+RM)      & 8B & 71.13 & 63.00 & 79.00 & 85.19\\

\bottomrule
\end{tabular*}
\end{table*}

\begin{figure}[t]
    \centering
    \includegraphics[width=1.0\linewidth]{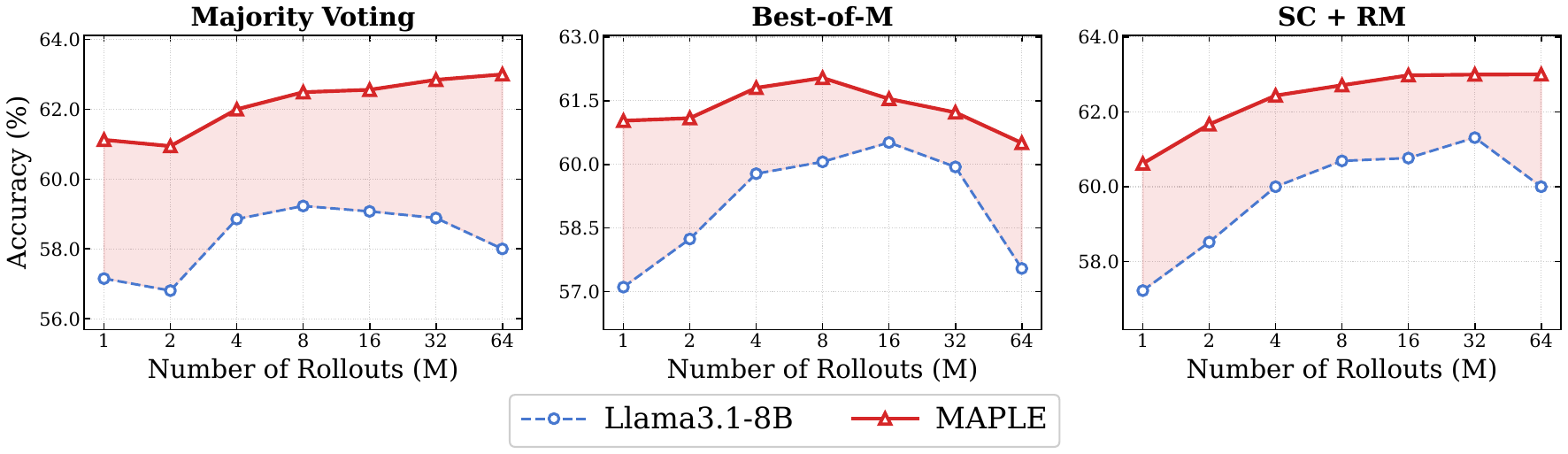}
    \caption{
Test-time scaling curves on MedMCQA under
: MV, BoM, and SC+RM. MAPLE (red) consistently outperforms the Llama3.1-8B backbone (blue) across different rollout budgets M
, with the shaded region highlighting the performance gap.
    }
    \label{fig:ablation}
\end{figure}


\noindent\textbf{Effect of MedPRM.} As shown in Table~\ref{tab:ablation}, comparing TTRL (69.84\% on MedQA) against
MAPLE (BoM) (73.02\%), incorporating MedPRM-guided reward during test-time optimization yields consistent improvements across all benchmarks, confirming that process-level medical reward is essential. 


\noindent\textbf{Test-time Scaling Behavior.} Fig.~\ref{fig:ablation} shows that as rollouts $M$ increase from 1 to 64, MAPLE consistently outperforms the Llama3.1-8B backbone under all decoding strategies, indicating robust gains across inference budgets. Accuracy for both models rises with $M$ and plateaus around $M=16$--$32$, after which improvements diminish and BoM may slightly degrade, likely due to reward model noise at large $M$. Notably, the MAPLE--backbone gap widens at larger $M$ under MV and SC+RM, suggesting MAPLE produces higher-quality, more diverse reasoning chains that benefit more from aggregation.

\section{Conclusion}
We propose MAPLE, a test-time training method for medical reasoning that replaces MV pseudo-supervision with step-wise rewards from a medical process reward model, delivering consistent and robust gains across rollout budgets on four medical QA benchmarks. MAPLE achieves state-of-the-art performance among 8B models, outperforming its Llama3.1 backbone, reasoning-distilled and PRM-only baselines, and even surpassing QwQ (32B) on DDXPlus and MMLU-Med despite being 4$\times$ smaller in parameter scale.
\bibliographystyle{splncs04}
\bibliography{mybibliography}
%




\end{document}